# Personalized Prediction Models for Changes in Knee Pain among Patients with Osteoarthritis Participating in Supervised Exercise and Education

M. Rafiei[1,2*], S. Das[1,2], M. Bakhtiari[1], E.M. Roos[3], S.T. Skou[3,4], D.T. Grønne[3,4], J. Baumbach[1,5], L. Baumbach[2]

1. Institute for Computational Systems Biology, University of Hamburg, Hamburg, Germany
2. Department of Health Economics and Health Services Research, University Medical Center Hamburg-Eppendorf, Germany
3. Center for Muscle and Joint Health, Department of Sports Science and Clinical Biomechanics, University of Southern Denmark, Odense, Denmark
4. The Research and Implementation Unit PROgrez, Department of Physiotherapy and Occupational Therapy, Næstved-Slagelse-Ringsted Hospitals, Slagelse, Denmark
5. Computational Biomedicine Lab, Department of Mathematics and Computer Science, University of Southern Denmark, Odense, Denmark

**Abstract**

**Background:**

Knee osteoarthritis (OA) is a widespread chronic condition that impairs mobility and diminishes quality of life. Despite the proven benefits of exercise therapy and patient education in managing the OA symptoms pain and functional limitations, these strategies are often underutilized. To motivate and enhance patient engagement personalized outcome prediction models can be utilized. However, the accuracy of existing models in predicting changes in knee pain outcomes remains insufficiently examined.

**Objective:**

To validate existing models and introduce a concise personalized model predicting changes in knee pain from before to after participating in a supervised patient education and exercise therapy program (GLA:D®) among patients with knee OA.

**Method:**

Our prediction models leverage self-reported patient information and functional measures. To refine the number of variables, we evaluated the variable importance and applied clinical reasoning. We trained random forest regression models and compared the rate of true predictions of our models with those utilizing average values. In supplementary analyses, we additionally considered recently added variables to the GLA:D® registry.

**Result:**

We evaluated the performance of a full, continuous, and concise model including all 34, all eleven continuous, and the six most predictive variables respectively. All three models performed similarly and were comparable to the existing model, with R-squares of 0.31-0.32 and RMSEs of 18.65-18.85 – despite our increased sample size. Allowing a deviation of 15 VAS points from the true change in pain, our concise model and utilizing the average values estimated the change in pain at 58% and 51% correctly, respectively. Our supplementary analysis led to similar outcomes.

**Conclusion:**

Our concise personalized prediction model provides more often accurate predictions for changes in knee pain after the GLA:D® program than utilizing average pain improvement values. Neither the increase in sample size nor the inclusion of additional variables improved previous models. To improve predictions, variables beyond those identified in the literature and collected as part of GLA:D® are required.

## Introduction

### Background

OA is a chronic degenerative disease that significantly affects the quality of life due to persistent pain and limited mobility, especially for those suffering from knee OA [1–3]. In the United States alone, the overall economic burden of OA is estimated at almost $140 billion annually, highlighting the substantial societal and personal costs associated with the condition [4]. In managing OA, patient education and exercise therapy, which are the recommended first-line treatments according to clinical guidelines play a critical role [5–7]. Exercise therapy, when consistently practiced, has been shown to improve pain, function, and quality of life [8]. Meanwhile, patient education empowers individuals to effectively manage their symptoms by deepening their understanding of the disease, its progression, and the significance of lifestyle adjustments like a balanced diet and regular exercise [9].

### Prior Work

However, implementing these recommended treatments could be improved [10–12]. The discrepancy between recommended and provided treatments has been attributed to several factors, including healthcare providers offering no evidence-based care and patients' lack of motivation and awareness about the long-term benefits of the recommended therapies [13,14]. Furthermore, some patients are hesitant to participate in exercise therapy programs, either because they are uninterested or face logistical difficulties [15,16]. It is relevant to develop strategies that assist clinicians and patients in an engaging way in the treatment decision process such as offering personalized outcome predictions.

Digital technology presents a promising solution to address such challenges in healthcare. Prediction models, particularly, have emerged as powerful tools to support patients and healthcare professionals in the treatment decisions and management of chronic diseases like OA [17–21].

**The goal of This Work**

The GLA:D® (Good Life with osteoArthritis in Denmark) program has garnered significant attention as a pioneering initiative in OA management [22]. Comprising three components – training for physiotherapists, a patient education and neuromuscular exercise therapy program delivered to patients, and clinical data collection – the program stands out as an example of evidence-based care [22]. To support shared decision-making, the acceptance, and participation rate of programs like GLA:D®, we aimed to validate an existing model and introduce an updated concise personalized prediction model that estimates changes in knee pain intensity for patients with OA considering participation in the GLA:D® program [23].

## Methods

This paper follows the guidelines of "Transparent Reporting of a Multivariable Prediction Model for Individual Prognosis or Diagnosis" (TRIPOD) [24].

In this paper, we reproduced and validated the results of previous work [23] with some methodological changes as described below.

**Source of Data and Participants**

We utilized data from the Danish GLA:D® initiative for knee and hip OA patients. The initiative consists of a 2-day course for physiotherapists, a patient treatment program delivered in clinical practice, and a registry of data reported by patients and clinicians. The patient program combines two patient education and 12 supervised exercise therapy sessions aiming at improving physical function and quality of life. The GLA:D® registry collects data from participants, which includes demographic details, medical history, pain intensity, and physical function measures [25]. The data is collected at three-time points: at baseline, immediately following the program, and at 12 months follow-up. GLA:D® was established in 2013 and is regularly updated to include new evidence. Comprehensive details about GLA:D®'s education, neuromuscular exercise program, and general information are available in other sources [22]. For the present paper, we included patients, who indicated their knee as the joint of primary complaint and provided complete data. This is in line with the previous publication as the exclusion of any participant who started between the 23rd of May 2016 and the 12th of November 2016, since one variable was not collected during this period, due to technical problems in the registry [23]. However, our inclusion period was longer involving patient information from those providing data between October 9th, 2014, and November 12th, 2022. Further details on the inclusion and exclusion process are provided in a flowchart in Figure 2.

**Outcome**

"Change in pain intensity" was our outcome of interest. This measure was calculated by determining the changes in pain scores from baseline to immediately after the program (after about 3 months). The question asked to the patients was "I would like you to think about a scale that goes from no pain (0) to worst pain imaginable (100) that best represents your knee pain during the last week." Participants indicated their pain level on a visual analog scale (VAS) ranging from 0 to 100 mm, where 0 represented "no pain" and 100 signified "worst possible pain" [26].

**Predictor variable selection**

Our analysis incorporated 34 potential predictor variables: 11 continuous, 22 binary, and one with three categories (Table 1 in Multimedia Appendix 1). These variables were chosen for their relevance to knee OA and health outcomes, encompassing factors like clinical symptoms, lifestyle influences, and demographic details, to provide a comprehensive understanding of the disease's impact and patient outcomes. In comparison to the earlier study, we included 17 fewer variables (Diabetes, S12 physical component and mental score, ASES pain and other symptoms score, and some medical disease) since their collection stopped (for four variables in 2018, for two variables in 2020, and for 12 variables their collection stopped in 2021) [23].

**Statistical analysis methods**

We utilized random forest regressions during variable selection and model development. The random forest regressor is an ensemble model that efficiently reduces variables without overfitting and captures complex, non-linear relationships between predictors and the outcome variable [27–29]. Random forest regression was previously used on the GLA:D® dataset to predict changes in VAS pain considering 51 variables in the training process; its performance is similar to a linear regression [23].

*Data preprocessing*

The preprocessing of our dataset was performed according to the previous study which results we aim to validate [23].

*Selection of variables*

We applied a two-step variable selection process to optimize our predictive model and attempt to achieve clinical acceptance. Initially, we started with 34 predictor variables. To reduce the number of variables, we first evaluated the variable importance using gini impurity [30]. The Gini Index or Impurity measures the probability of a random instance being misclassified when chosen randomly. The Gini impurity method evaluates the importance of each variable by examining how much it contributes to reducing impurity when used as a split criterion in decision trees [31]. Variables with higher Gini importance scores are considered more important in making accurate predictions. In many implementations, like scikit-learn's random forest, the Gini importance scores are normalized such that they all sum up to 1. Therefore, each variable's importance score will range between 0 and 1, with higher values indicating greater importance. We built and trained random forest regressor models, incorporating the top k variables, with k determined using the elbow method. The elbow method is a simple way to identify the optimal number of variables in a model, by finding the point where adding more variables brings little to no additional benefit [32]. To enhance the robustness of our analysis, we employed cross-validation during the variable importance assessment.

The Gini coefficient is mathematically represented as:

$$Gini = 1 - \sum_{i=1}^{C} (p_i)^2$$

In the second step, we respected the importance of the individual variables and the elbow technique additionally we applied clinical reasoning to reduce and find the ideal set of variables. While this reduction of variables could potentially reduce the model's performance, we limited the number of questions, such that it would be convenient for end-users to utilize the model, thereby increasing the likelihood of its clinical value.

With this two-step approach, incorporating both Gini Impurity and clinical reasoning, we aimed to achieve a balanced model that is both predictive and user-friendly.

### *Model development and evaluation*

To develop and evaluate our predictive model, we began by splitting our dataset into two subsets: a training set for model development and a test set for validation. The training process was consistently applied on all models, regardless of training on all variables, on the top k predictor variables, or on a concise set of variables. To enhance the robustness and generalizability of our models, we integrated 10-fold cross-validation with random forest regression. Utilizing the test sets, we validated our findings, ensuring that our results were not biased or limited to specific data instances by averaging results across the cross-validation test sets.

To evaluate the performance of the models, we used the root mean squared error (RMSE) and R-squared (R2) metrics. RMSE measures the average deviation between the predicted and actual values, providing an assessment of the model's accuracy. The R2 represents the proportion of variance the model explains and indicates how well the model fits the data. To compare our personalized outcome predictions with predictions of the average model we evaluated the number of correct predictions, in contrast to the previous publication, which used the absolute mean differences in a test data set [23].

The predicted changes in pain intensity by our random forest regression constitute the personalized predictions. The average model prediction was defined as a mean function of the VAS pain change score of the training samples, $\mu = \frac{\sum_{n=1}^{N} VASChangeScore}{N}$. The average model always provides μ, regardless of the features of the unseen samples, as the predicted change in pain intensity. In this paper, we are always focused on the VAS pain change score, rather than the pain score at a specific time point. Recognizing that predictions inherently have some margin of error, we incorporated in our evaluation a margin of error that corresponds to a clinically relevant threshold.

$$I_n = \begin{cases} 1: & \text{if } L_n \leq \text{Pred}_n \leq U_n \\ 0: & \text{Otherwise} \end{cases}$$

$I_n$ is the indicator value that will be one for cases where their predicted change in pain value is inside the interval. For all other samples, it is zero. $L_n = VASChangeScore - R$ and $U_n = VASChangeScore + R$ show the lower and upper bound based on the clinical relevance threshold R. Finally, we calculate the percentage of predictions inside the interval ρ to evaluate and compare our model against the average model:

$$\rho = \frac{\sum_{n=1}^{N} I_n}{N}.$$

Since the value of a clinical relevance R is debatable and ranges up to 20 mm [33–35] points we evaluated the number of correct predictions with different tolerable margins of error, ranging from 5 to 20 points. Thus, we calculated the number of correct predictions for several cases, the first allowing a deviation of 5 points and the last allowing a deviation of 20 points from the true change in pain.

All data analyses were performed in Python using Scikit-learn, Pandas, and Numpy libraries. We utilized Pandas and Numpy for data preprocessing tasks, such as cleaning, and variable scaling. Scikit-learn was utilized for validation techniques, including train-test split, cross-validation, and stratified sampling. We used standard hyperparameters for the random forest regression: n_estimators set to 100, max_depth limited to 10, and random_state fixed at 42. We chose to use them to provide a baseline performance that can be compared with other standard implementations.

*Supplementary analyses*

To evaluate if the inclusion of recently added variables to the GLA:D® registry improved our outcome predictions, we additionally developed and evaluated models following the same process as described above, but including 12 additional variables (Load-related pain,

Reduced functional capacity, Morning stiffness, Crepitus, Reduced knee movement, Bony enlargement, Previous joint injury, Occupational or recreational overuse, Family members with OA, KOOS-12 pain, Functioning and KOOS-12 summary score, which were added to the registry in May 2018). A flow chart for this additional analysis is provided in Figure 2 in Multimedia Appendix 1.

## Results

A brief overview of our results is displayed in Figure 1.

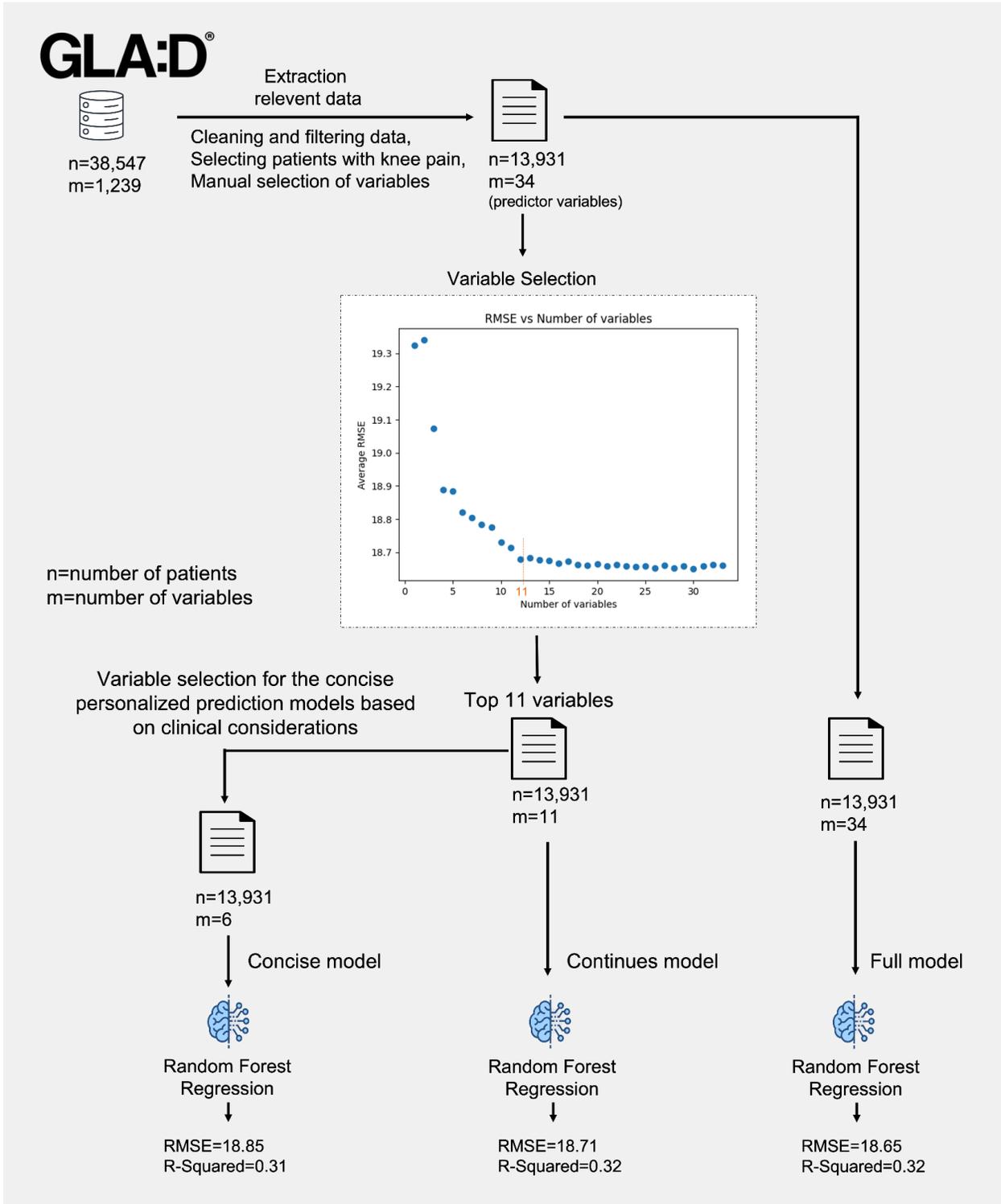

Figure 1. A brief overview of the personalized prediction to predict changes in knee pain in patients with knee OA

**Participants**

38,547 patients fulfilled our inclusion criteria for participating in the GLA:D® program between October 9th, 2014, and November 12, 2022. We considered patients with knee pain the primary complaint and our analytical dataset was reduced to 13,931 patients. Details on the process of patient inclusion and exclusion are available in Figure 2.

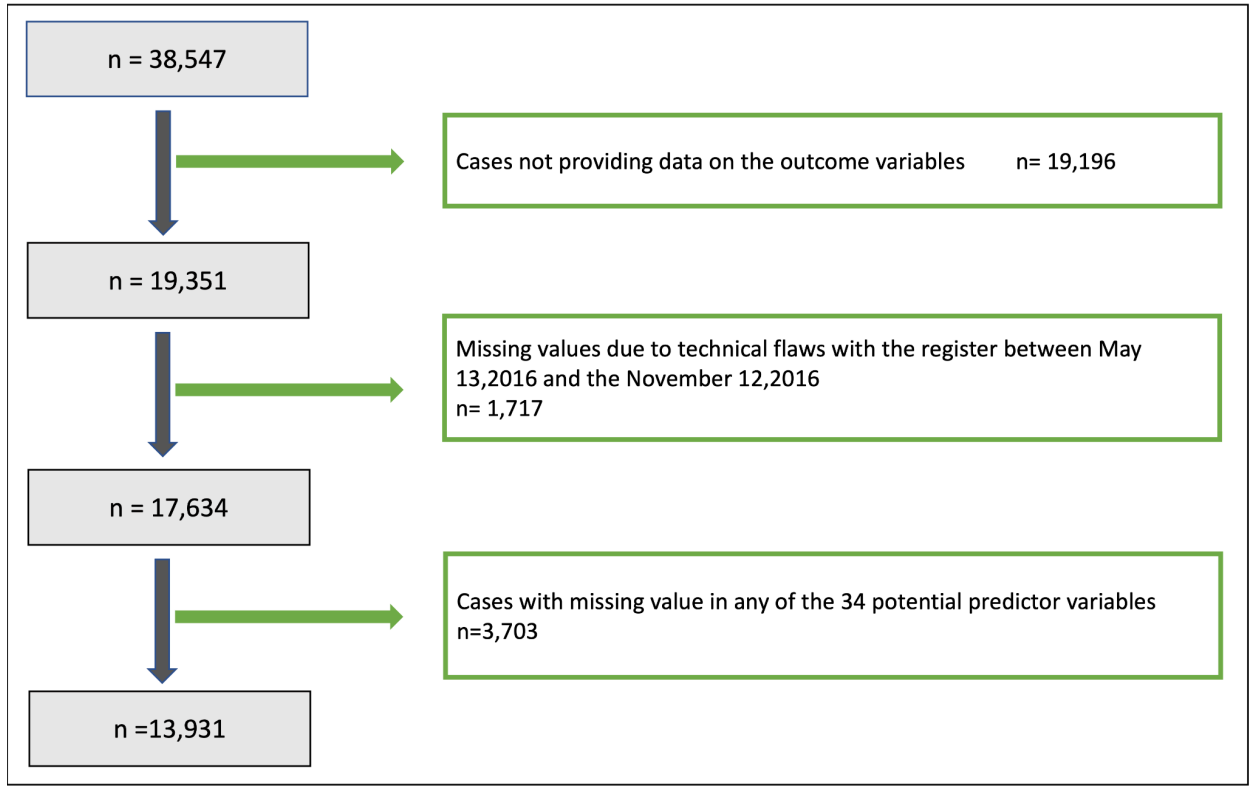

Figure 2. Flowchart of patient selection and data exclusion criteria for the study, showing initial cohort size and subsequent exclusions due to specific criteria and missing data

**Variable Selection and Model Development**

The Gini Impurity scores for all variables are presented in Figure 1 in Multimedia Appendix 1. Based on these findings, we applied the Elbow method and included the top 11 variables (Figure 3). In total, we built 3 models each including a different set of variables (Table 1). The first includes all 34 variables (Full model). The second one includes the 11 most important variables identified with the Gini Impurity and Elbow method (Continuous model), highlighting the importance of these baseline variables in making predictions. The third includes the 6 predictive variables, which were chosen based on their importance and clinical reasoning – it is assumed that answering the underlying 11 questions would be applicable in routine clinical practice. These six variables are Baseline pain, Duration of Symptoms, the EQ-5D Score, Time to Complete a 40m Walking Test, Age, and BMI. Each of those 6 variables contains one item but BMI and EQ-5D which include 2 and 5 respectively.

| Name of Variable combination | Included variables |
|---|---|
| 1. Full model | All variables |
| 2. Continues model | The 11 most important variables were selected by Gini. identified by variable selection |
| 3. Concise model for clinical practice | Age, BMI, Change in pain, Duration of symptoms, Time to complete 40m walking test, EQ-5D score |

BMI= Body Mass Index

Table 1: Overview of included variables per model

Gini Impurity to identify the most important variables:

| | Variable | Importance |
|---|---|---|
| 1 | Pain intensity in the index joint during the last month/week (VAS scale 0-100, no pain to worst pain) | 0,530013 |
| 2 | EQ-5D score | 0,065382 |
| 3 | Time to complete 40m walking test | 0,05466 |
| 4 | Duration of symptoms | 0,054654 |
| 5 | BMI | 0,043376 |
| 6 | EQ VAS general health (0-100, worst to best) | 0,040781 |
| 7 | Age | 0,037102 |
| 8 | KOOS-12 Quality of life subscale score | 0,028261 |
| 9 | Number of chair stands during 30sec | 0,025484 |
| 10 | Number of painful body areas (collected via pain manneqin, 0-56) | 0,022765 |
| 11 | UCLA - physical activity score (from 0-10, worst to best) | 0,017916 |
| 12 | Are your hip/knee problems so severe that you would like an operation? | 0,009908 |
| 13 | Educational level Do you have an education higher than secondary education? | 0,005545 |
| ... | ... | ... |
| 33 | Danish citizen | 0,00077 |
| 34 | Pain in other hip or knee joints than the index joint? | 0,000016 |

Elbow method to find top 11 variables:

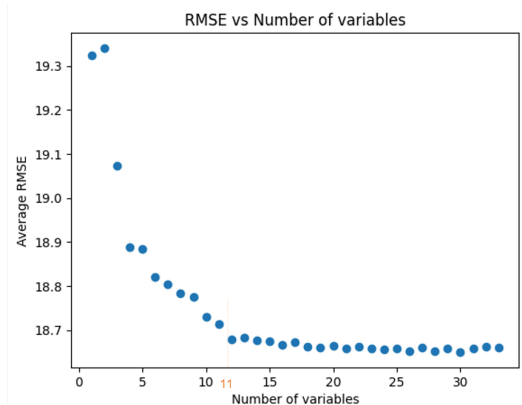

Figure 3. Demonstrates the two-fold analytical process where the left panel presents the ranking of variables by their relative importance using Gini Impurity. Following this initial analysis, the Elbow method, showcased in the right panel, was applied.

**Model evaluation**

The model performances are presented in Table 2 based on the average performance on the test sets in a 10-fold cross-validation process. Accordingly, the full model's average Root Mean Square Error (RMSE) was 18.65, while the coefficient of determination (R-squared) was 0.32. The continuous model displayed an RMSE of 18.71 and an R-squared of 0.32. Moreover, when considering variables in the concise model, the RMSE and R-squared were 18.85 and 0.31, respectively.

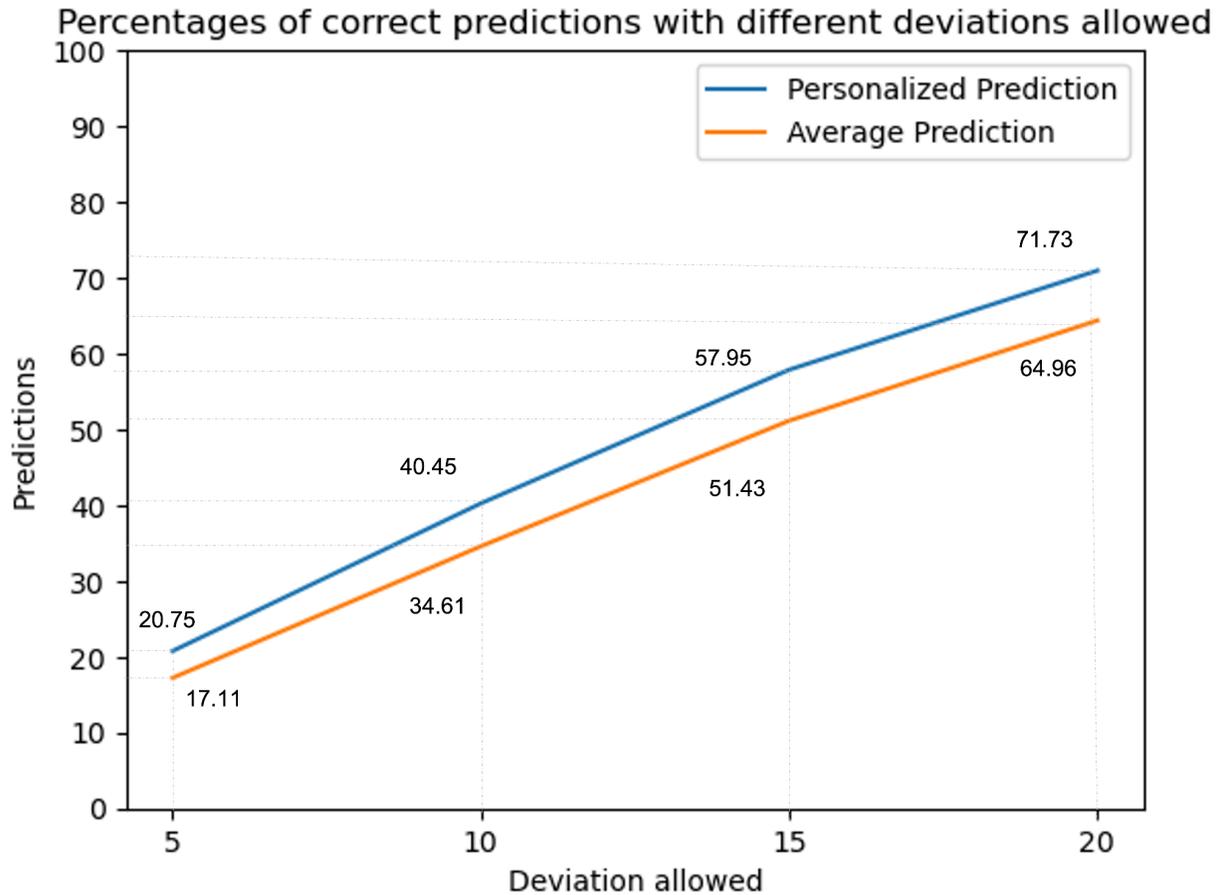

Figure 4: Comparison of the percentage of correct predictions of our concise model and utilizing the average model, allowing deviations between 5 and 20 points

The comparison between true predictions based on our concise model including the top 6 most important variables and predictions using the average model is illustrated in Figure 4. It shows a small but distinct advantage of our personalized predictions over average-based predictions. Our personalized prediction method, based on random forest regression, performed better than the average prediction model by a margin of about 7% when allowing a difference of 15 points in each direction between the predicted and the true change on the test data [33]. Specifically, our results showed that the average model predicts 51.43% and our personalized model predicts about 57.95% of the cases correctly. We found that personalized predictions were slightly and consistently better than those based on the average improvement, regardless of the deviation allowed (5 points or 20 points). Table 2 shows correct personalized predictions and correct average predictions within the interval of +/-15 points for all models. An illustration utilizing these numerical values is as follows: If a patient, for instance, has a pain score of 45 before GLA:D® and our personalized model estimates a change of 20 points, this means that it is estimated that the patient will have pain between 10 and 40 after the program with a 58% certainty.

| Variable combination | VAS Pain: GLA:D® Personalized prediction | | | |
|---|---|---|---|---|
| | RMSE | R-Squared | Correct personalized predictions (within the interval of +/- 15 points) | Correct average predictions (within the interval of +/- 15 points) |
| 1. Full model | 18.65 | 0.32 | 58.38% | 51.43% |
| 2. Continues model | 18.71 | 0.32 | 58.28% | 51.43% |
| 3. Concise model for clinical practice | 18.85 | 0.31 | 57.95% | 51.43% |

**RMSE= Root Mean Squared Error, VAS = Visual Analog Scale.**

Table 2: Overview of the performance of all random forest regression models

### Supplementary analyses

Including the 12 additional variables reduced our sample with complete cases to n= 4,908, since these variables were first added later. However, the evaluation of the variable importance followed a similar pattern highlighting that the now 14 continuous variables were the most important (Figure 2 in Multimedia Appendix 1). Similarly, the model performances and evaluations remained comparable among all three models (Table 3).

| Variable combination | VAS Pain: GLA:D® Personalized prediction | | | |
|---|---|---|---|---|
| | RMSE | R-Squared | Correct personalized predictions (within the interval of +/- 15 points) | Correct average predictions (within the interval of +/- 15 points) |
| 1. Full model | 18.96 | 0.32 | 58.35% | 51.14% |
| 2. Continues model | 19.04 | 0.32 | 58.41% | 51.14% |
| 3. Concise model for clinical practice | 19.21 | 0.31 | 57.86% | 51.14% |

**RMSE= Root Mean Squared Error, VAS = Visual Analog Scale.**

Table 3: Overview of performance summary of all random forest regression models of our supplementary analyses, incorporating 46 variables with 12 additional variables relative to the prior model. This includes 30 binary variables, 2 categorical variables, and 13 continuous variables.

**Discussion**

In this study, we developed personalized prediction models for changes in knee OA pain after supervised patient education and exercise therapy (GLA:D®) and compared them to existing and average-based models. Our findings validate the previously developed models [23] and introduce a concise personalized prediction model for estimating changes in knee pain intensity among patients with knee OA considering participating in the GLA:D® program. Neither the increase in sample size nor the inclusion of additional variables improved the previous model. Nonetheless, our concise model correctly predicted 58% of the cases and demonstrated a 7% improvement in the rate of correct predictions over the use of currently used average values for informing patients about their expected changes in knee pain.

In clinical decision-making, predictive models serve as pivotal tools across various disciplines. These models, akin to those used in cancer prognosis or the optimization of exercise regimens, underline the significance of personalized predictions in enhancing patient outcomes [36–40]. Our study delves into this paradigm by comparing prediction models and validating the findings of an earlier study. As in the previous study [23], we utilized the GLA:D® data and presented a full and continuous model to predict personalized outcomes. The latter was chosen based on the variable importance. While we utilized the Gini impurity and elbow method, the previous study determined the most important variables by the reduction in Root Mean Square Error (RMSE) for out-of-bag cases and applying the elbow method [23]. Despite the difference in the utilized method, our findings that the continuous variables were most important align with the previous study. Our outcomes match as well those of the previous study despite our model's sample size being doubled and the variables being reduced by 17 and 4 in the full and continuous model, respectively. For the full and continuous models, there was no difference, and a difference of 0.01 in the R2s, respectively. Also, the RMSE aligned closely. In our supplementary analysis, we incorporated 12 new variables, included in the GLA:D® registry between 2017 and 2018, which reduced our sample size to 4,908 but maintained consistent variable importance and model performance. Thus we found that neither an increase in sample size nor the added variables could improve the personalized outcome predictions compared to the earlier study [23].

To compare the personalized predictions with predictions based on average values, we again utilized a different approach than the previous study. The previous study [23] took the absolute mean difference and concluded that the difference between their model and the average improvements was not clinically relevant. In this study, we focused on the number of correct predictions, assessing a range of clinical relevance thresholds to determine whether our personalized model surpasses the average model's performance, despite changing the clinical relevance threshold. Here, we emphasize this incremental progress of increasing the clinical relevance threshold. As a result, allowing a higher deviation from the true change resulted in a higher number of correct predictions, for both our personalized predictions and the average prediction. Overall we found that our personalized prediction model was able to predict about 7% more cases correctly compared to utilizing average predictions, independent of the threshold allowed from the true change.

However, the question of whether these findings are sufficient for implementation into clinical practice remains open. Existing shared decision-making tools often concentrate on comparing different treatments and assisting patients in making informed choices [20]. Specifically, in the context of OA, where education and exercises are universally recommended and surgery is considered a last resort after the failure of previous treatments, the potential of improved prediction models to alter this recommendation landscape is significant [41]. If future models could reliably predict the likelihood of initial treatment failures for specific patient groups, clinical guidelines to treat OA could evolve. Yet, it's crucial to acknowledge that the current performance of our models is too preliminary for such decisive changes in clinical practice. This underscores the necessity for further research to enhance model accuracy and reliability as well as to understand the needs and preferences of patients and clinicians, potentially reshaping treatment protocols for OA based on predictive insights.

**Limitation**

Our study also has some limitations. Selection bias is possible in any registry-based study due to the cases lost to follow-up. However, there was only a 28.44% loss to follow-up in the study, which suggests no serious threats to its external validity [42]. Our study suggests that the increased sample size does not significantly affect the predictive performance outcome [23]. However, it should be noted that we included fewer variables in comparison to the previous study [23]. These variables were deemed of less relevance and therefore excluded from the registry, nonetheless, they contributed to the predictions of the previous models. Therefore, to be precise, we can only conclude that the increased sample size was able to compensate for the reduced number of variables. Hence, to improve our models' predictability, new variables would probably need to be integrated. The newly integrated variables (12 additional variables) in our supplementary analyses, led to a reduction in the sample size (n=4,908), however, the model performance remained similar. Consequently, we can (only) conclude that the newly added variables could compensate for the reduced sample size.

Based on these observations we can point toward the limitation of available predictor variables. Implementing more detailed questionnaire-based measures for anxiety and depression might enhance pain predictions [43]. Weight loss is another recommended first-line treatment for overweight and obese individuals with knee OA [44] that is known to moderate systemic inflammation [45], extending the questionnaire to enquire about patients' dietary habits might thus be another possibility to increase the predictive value [46]. However, due to the clinical applicability, the extent of the GLA:D® registry questionnaire needs to balance the collection of information against patient burden, adding additional variables with minor importance is not desired. In conclusion, we currently do not know which variables could have an additional major positive influence on the personalized outcome predictions over those already included [47].

**Future Directions & Considerations**

The precision of our model aligns with existing clinical studies, suggesting a need for future investigations into the practical value of such predictive accuracy within clinical settings [48]. Current initiatives like the A.S.K. report and Movement is Life™ have begun to explore the integration of predictive models and patient-reported outcomes to guide shared decision-making, demonstrating the utility of these tools in clinical practice [20,49]. However, there is a noticeable gap in the literature concerning the specific performance benchmarks required for predictive models in physiotherapy to be deemed clinically relevant, particularly in the management of OA. Therefore, before a decision on implementing our model as an online tool into clinical practice can be drawn, guidance on which metrics and predictive values need to be achieved in a model to be feasible for clinical practice needs to be investigated. Furthermore, it remains uncertain if the implementation of a model, like ours, which provides more correct predictions than what is currently used, but still several wrong predictions, is valuable enough for patients and clinicians to be implemented in clinical practice. Further to the best of our knowledge, it is yet unknown if personalized outcome predictions regarding a treatment lead to different expectations than average outcome predictions in patients. Moreover, on ethical aspects, the consequences of wrong predictions in general on a patient's treatment, and mental and emotional well-being should be investigated. For patients with estimated worsening pain predicting other health outcomes additionally could be essential to motivate them to participate in the program. For example prediction tools within GLA:D® for secondary outcomes such as physical activity, which changes independently of changes in pain, could support participation rates, given the overarching health benefits associated with increased physical activity [25,47]. Similarly, patients might be more motivated if they knew about the expected changes in physical function and quality of life. Finally, any prediction tool needs to be comprehensively evaluated based on its role in treatment decisions, patient satisfaction, and overall treatment outcomes in clinical settings, before implementation.

**Generalisability**

The methodology employed in our study, which leverages machine learning techniques to analyze questionnaire data and functional test data for predicting treatment outcomes has been utilized before and is a promising advancement in the field of personalized medicine [50,51]. This approach is characterized by the selection of predictive variables from data and can be utilized in similar questioning. Thus, our method extends beyond the specific context of the GLA:D® program and could be applied to other exercise therapy programs targeting knee OA, and potentially, for a broader range of conditions [49,52–54].

Generalizing our findings, the model predictions of the GLA:D® program for knee OA to other exercise and physiotherapy interventions are challenging. The unique combination of group-based exercises and education in GLA:D® may be crucial and not present in all physiotherapeutic settings. Therefore, our outcome predictions specific to GLA:D® cannot be expected universally. However, if the core principle of structured exercise therapy and education are incorporated into the treatment of a patient with OA, our model might be applicable. Nonetheless, the personalized predictions would likely deviate even further from the true changes than in patients participating at GLA:D®, therefore they shouldn't be generalized.

## Conclusion

In conclusion, we developed a concise personalized prediction model that estimates individual changes in pain intensity for patients with knee OA considering participation in the GLA:D® program. The variance explained by our model is limited and was not improved with larger sample sizes or additional variables. To improve predictions, variables beyond those identified in the literature and collected as part of GLA:D® are required. Nonetheless, our personalized model predicts 58% correctly, and 7% of patients more correctly than utilizing average values. However, whether the performance is good enough for clinical practice is questionable. Guidance under which circumstances and how well a model needs to perform to be implemented in clinical practice is therefore needed.

## Availability of data and material

All the data for this study were acquired following a request to the GLA:D® registry, details of which can be found on their official website: https://www.glaid.dk/index.html.

## Funding sources

This study was developed as part of the PhysioAI project and is funded by the German Federal Ministry of Education and Research (BMBF) and by the NextGenerationEU Fund of the European Union under grant number 16DKWN115A. The initiation of GLA:D® was partly funded by the Danish Physiotherapy Association's fund for research, education, and practice development; the Danish Rheumatism Association; and the Physiotherapy Practice Foundation. The responsibility for the content of this publication lies with the author.

## Acknowledgments

The authors would like to thank the clinicians and patients collecting data for the GLA:D® registry.

## Disclosure of interest

EMR is on the Editorial Board of Osteoarthritis and Cartilage, the developer of the Knee Injury and Osteoarthritis Outcome Score (KOOS) and several other freely available patient-reported outcome measures and co-founder of GLA:D®, a not-for-profit initiative hosted at the University of Southern Denmark aimed at implementing clinical guidelines for osteoarthritis in clinical practice. STS has received personal fees from Munksgaard, TrustMe-Ed, and Nestlé Health Science, outside the submitted work, and is co-founder of GLA:D®.

## Abbreviations

OA: osteoarthritis

## Multimedia Appendix 1

Additional information about this study can be found online at [[Link](Link)].

**Multimedia Appendix 1:**

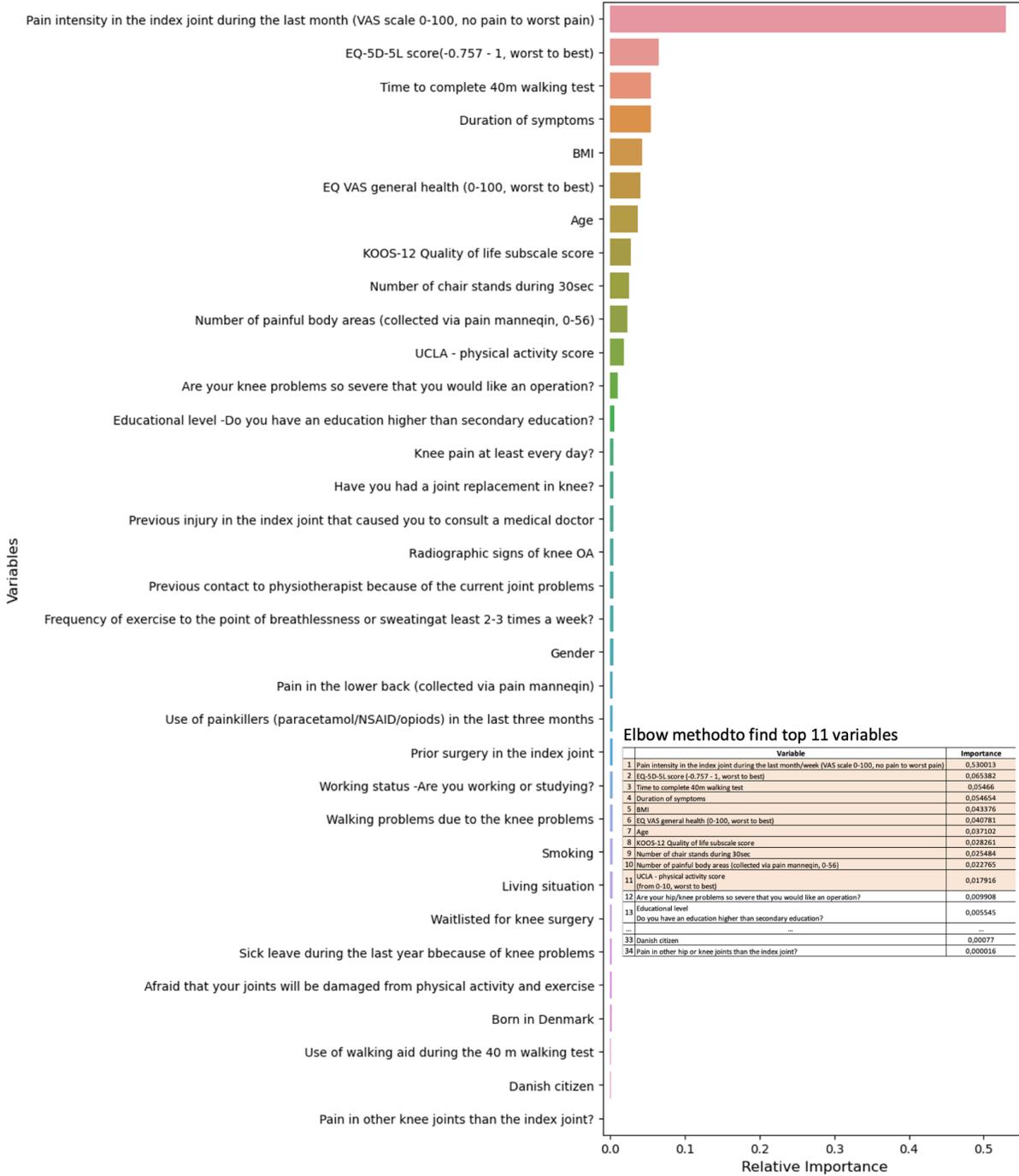

Figure 1: Visual breakdown of the top variables affecting VAS pain change scores, ranked by a random forest regressor using Gini Impurity from a 34 variable, with an inset highlighting the 11 most critical variables as determined by the elbow method.

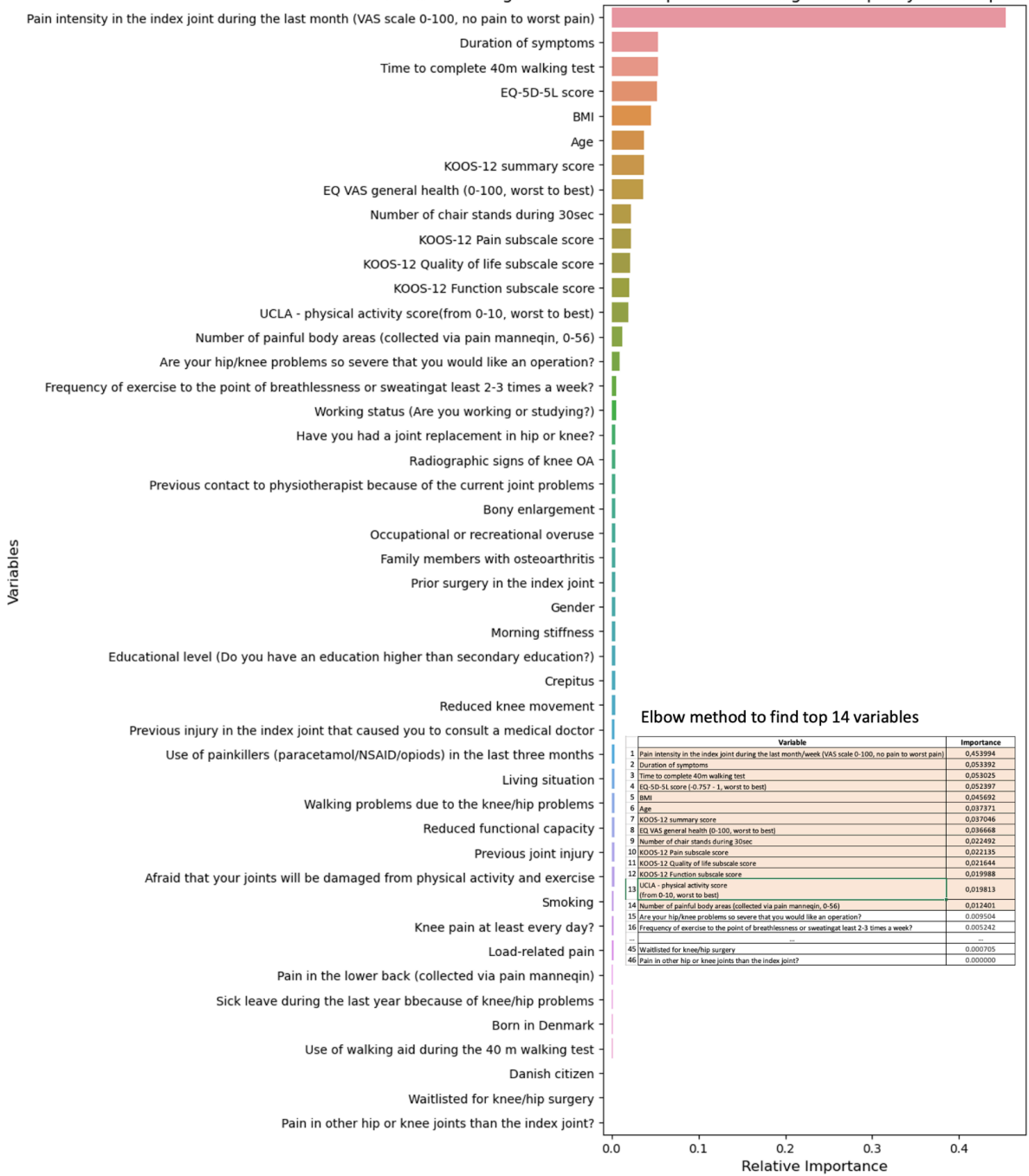

Figure 2: Visual breakdown of the top variables affecting VAS pain change scores, ranked by a random forest regressor using Gini Impurity from a 46-variable, with an inset highlighting the 14 most critical variables as determined by the elbow method.

| Variable | Included cases Mean(sd) or number of cases (%) n =13931 | Type of variable |
|---|---|---|
| *predictor variables* | | |
| 1. Age | Mean: 65.33<br>SD: 9.28<br>Min: 23<br>Max: 94 | Continuous |
| 2. BMI | Mean: 28.65<br>SD: 5.29<br>Min: 15.23<br>Max: 70.03 | Continuous |
| 3. Gender | Man: 4085<br>Woman: 9846 | Categorical (2) |
| 4. Duration of symptoms | Mean: 40.54<br>SD: 64.10<br>Min: 0.00<br>Max: 756.0 | Continuous |
| 5. Waitlisted for knee/hip surgery | Yes: 207<br>No: 13724 | Binary |
| 6. Radiographic signs of knee OA | Yes: 11079<br>No: 511<br>Unknown: 2341 | Categorical (3) |
| 7. Previous contact to physiotherapist because of the current joint problems | Yes: 4644<br>No: 9287 | Binary |
| 8. Use of painkillers (paracetamol/NSAID/opiods) in the last three months | Yes: 8809<br>No: 5122, | Binary |
| 9. Prior surgery in the index joint | Yes: 3790<br>No: 10141 | Binary |
| 10. Time to complete 40m walking test | Mean: 28.23<br>SD: 7.69<br>Min: 10.0<br>Max: 234.91 | Continuous |
| 11. Use of walking aid during the 40 m walking test | Yes: 229<br>No: 13702 | Binary |
| 12. Number of chair stands during 30sec | Mean: 11.97<br>SD: 3.70<br>Min: 0.00<br>Max: 40.00 | Continuous |
| 13. Born in Denmark | Yes: 13442<br>No: 489 | Binary |
| 14. Danish citizen | Yes: 13718<br>No: 213 | Binary |
| 15. Living situation | Yes: 3361<br>No: 10570 | Binary |
| 16. Educational level<br>Do you have an education higher than secondary education? | Yes: 9895<br>No: 4036 | Binary |
| 17. Smoking | Yes: 1066<br>No: 12865 | Binary |
| 18. Previous injury in the index joint that caused you to consult a medical doctor | Yes: 7063<br>No: 6868 | Binary |

| Variable | Included cases Mean(sd) or number of cases (%) n =13931 | Type of variable |
|---|---|---|
| 19. Pain in other hip or knee joints than the index joint? | Yes: 1<br>No: 13930 | Binary |
| 20. Walking problems due to the knee/hip problems | Yes: 10716<br>No: 3215 | Binary |
| 21. Knee pain at least every day? | Yes: 11214<br>No: 2717 | Binary |
| 22. Afraid that your joints will be damaged from physical activity and exercise | Yes: 2028<br>No: 11903 | Binary |
| 23. Pain intensity in the index joint during the last month (VAS scale 0-100, no pain to worst pain) | Mean: 46.40<br>SD: 21.66<br>Min: 0.00<br>Max: 100.00 | Continuous |
| 24. Number of painful body areas (collected via pain drawing) | Mean: 2.78<br>SD: 3.25<br>Min: 0.00<br>Max: 40.00 | Continuous |
| 25. Pain in the lower back (collected via pain manneqin) | Yes: 2690<br>No: 11241 | Binary |
| 26. Are your hip/knee problems so severe that you would like an operation? | Yes: 1549<br>No: 12382 | Binary |
| 27. Working status<br>Are you working or studying? | Yes: 4111<br>No: 9820 | Binary |
| 28. Sick leave during the last year bbecause of knee/hip problems | Yes: 1510<br>No: 12421 | Binary |
| 29. Have you had a joint replacement in hip or knee? | Yes: 1252<br>No: 12679 | Binary |
| 30. Frequency of exercise to the point of breathlessness or sweatingat least 2-3 times a week? | Yes: 6972<br>No: 6959 | Binary |
| 31. UCLA - physical activity score<br>(from 0-10 worst to best) | Mean: 5.72<br>SD: 1.77<br>Min: 1.00<br>Max: 10.00 | Continuous |
| 32.EQ VAS general health<br>(0-100, worst to best) | Mean: 70.53<br>SD: 18.54<br>Min: 0.00<br>Max: 100.00 | Continuous |
| 33. KOOS-12 Quality of life subscale score | Mean: 46.17<br>SD: 15.12<br>Min: 0.00<br>Max: 100.00 | Continuous |
| 34. EQ-5D-5L score<br>(-0.757 - 1, worst to best) | Mean: 0.78<br>SD: 0.18<br>Min: -0.50<br>Max: 1.00 | Continuous |
| *Outcomes variable* | | |
| 35. VAS pain change score from baseline to 3-month follow-up score<br>-100 -100, from pain, got worse to pain got better, 0 no change in pain | Mean: 14.06<br>SD: 22.75<br>Min: -88.0<br>Max: 99.0 | Continuous |

Table 1. Overview of registry questions for the included variable